\documentclass[letterpaper, 10 pt, conference]{ieeeconf}  

\IEEEoverridecommandlockouts                              

\usepackage{amsmath,amsfonts,amssymb}
\usepackage{booktabs}
\usepackage{graphicx}
\usepackage{cite}
\usepackage{multirow}
\usepackage{float}
\usepackage{color}
\usepackage{hyperref}
\hypersetup{
    colorlinks=false,
    pdfborder={0 0 0},
}

\DeclareMathOperator{\argmax}{arg\,max\,}
\DeclareMathOperator{\argmin}{arg\,min\,}

\title{\LARGE \bf
Cross-domain Transfer Learning and State Inference for Soft Robots via a Semi-supervised Sequential Variational Bayes Framework
}

\author{Shageenderan Sapai$^{1,*}$, Junn Yong Loo$^{1,*}$, Ze Yang Ding$^{2}$, Chee Pin Tan$^{2}$, Rapha\"{e}l C.-W. Phan$^{1}$, \\ Vishnu Monn Baskaran$^{1}$ and Surya Girinatha Nurzaman$^{2}$
\thanks{$^{*}$These authors contributed equally.}%
\thanks{$^{1}$The authors are with School of Information Technology, Monash University Malaysia 
(E-mail: shageenderan.sapai/loo.junnyong/raphael.phan/vishnu.monn@monash.edu).
}
\thanks{$^{2}$The authors are with School of Engineering, Monash University Malaysia 
(E-mail: ding.zeyang/tan.chee.pin/surya.nurzaman@monash.edu).
}
}

\begin{document}

\maketitle
\thispagestyle{empty}
\pagestyle{empty}

\begin{abstract}
Recently, data-driven models such as deep neural networks have shown to be promising tools for modelling and state inference in soft robots. However, voluminous amounts of data are necessary for deep models to perform effectively, which requires exhaustive and quality data collection, particularly of state labels. Consequently, obtaining labelled state data for soft robotic systems is challenged for various reasons, including difficulty in the sensorization of soft robots and the inconvenience of collecting data in unstructured environments. To address this challenge, in this paper, we propose a semi-supervised sequential variational Bayes (DSVB) framework for transfer learning and state inference in soft robots with missing state labels on certain robot configurations. Considering that soft robots may exhibit distinct dynamics under different robot configurations, a feature space transfer strategy is also incorporated to promote the adaptation of latent features across multiple configurations.
Unlike existing transfer learning approaches, our proposed DSVB employs a recurrent neural network to model the nonlinear dynamics and temporal coherence in soft robot data. The proposed framework is validated on multiple setup configurations of a pneumatic-based soft robot finger. Experimental results on four transfer scenarios demonstrate that DSVB performs effective transfer learning and accurate state inference amidst missing state labels. The data and code are available at \href{https://github.com/shageenderan/DSVB}{this GitHub repository}.
\end{abstract}


\section{INTRODUCTION}
Soft robots, often inspired by the morphology of bio-organisms \cite{lee2017soft}, are robots that can bend and conform to their surrounding structures \cite{laschi2016soft}. While this flexibility affords them advantages for deployment in unstructured environments \cite{kim2013soft}, it complicates the development of analytical models due to the inherent nonlinearities and theoretically infinite degrees of freedom \cite{nurzaman2013soft}. Conversely, data-driven methods present a more accessible alternative to soft robot modelling as the burden of describing the complex system dynamics is discovered from available data. For instance, \cite{bending_angle_pred, thuruthel_embedded_sensors_rnn, 3dsoftbody_pred} have successfully trained data-driven neural networks to address the complexity in modelling the state of soft robotic systems. Nonetheless, these data-driven approaches rely on the availability of high-quality labelled data, representative of the soft robots \cite{JUNN}. Such models demonstrate significant degradation in performance in the absence of labelled data \cite{soro_ml_review2}.


Unfortunately, data collection is often an arduous process as soft robots must be exhaustively actuated in order to extensively cover their task space \cite{soro_ml_review2}, \cite{rev-soroml}. This process is further challenged in some potential soft robot applications where collection processes are expensive \cite{octopus_robot, fish_robot, expensive_underwater_robots,medical_data_mining} due to underlying environmental conditions. In addition, soft robots exhibit distinct dynamics under different robot configurations \cite{thuruthel_embedded_sensors_rnn, JUNN} due to their inherent softness and compliance. Data would have to be collected on many configurations to accurately represent a versatile soft robot. Moreover, as opposed to the measurement inputs, collecting relevant state labels is non-trivial \cite{Zhuang1} and could give rise to missing state labels. For example, the installation of multiple sensors for capturing labelled state data is often avoided to maintain the soft robot's flexibility. Furthermore, external sensors such as cameras that track the morphological states of the robot are impractical in the unstructured environments within which these robots operate \cite{perceptive_soft_robots}. These challenges inhibit the deployment of data-driven models for soft robots.

To circumvent the requirement of voluminous data, previous studies \cite{bending_angle_pred, gaussian_process} employ simple or non-parametric models that are less reliant on data abundance but at the cost of having less expressive models. An alternative approach is to generate simulation data to compensate for the lack of real data by employing empirical simulation models. For example, \cite{runge2017fem, massari2020machine} generate simulated data for neural network training via the Finite Element Method (FEM). However, FEM models are generally sensitive to characterizing factors such as varying material properties, inconsistent manufacturing processes, and the influence of external loads \cite{fem}. A variation in these factors could result in substantial differences between the simulation models and the actual soft robots \cite{simulated_data_bad}. For this reason, data-driven models trained on simulated data often fail to generalize to highly uncertain real-world scenarios \cite{simulated_data_bad}. To overcome this, recent works \cite{Kriegman, Park, Donat, Wiese, Fang} incorporate sim-to-real transfer strategies to eliminate the gap between the FEM-simulated data and the physical experiment.

\begin{figure*}[tb]
\centering
\includegraphics[width=1\textwidth]{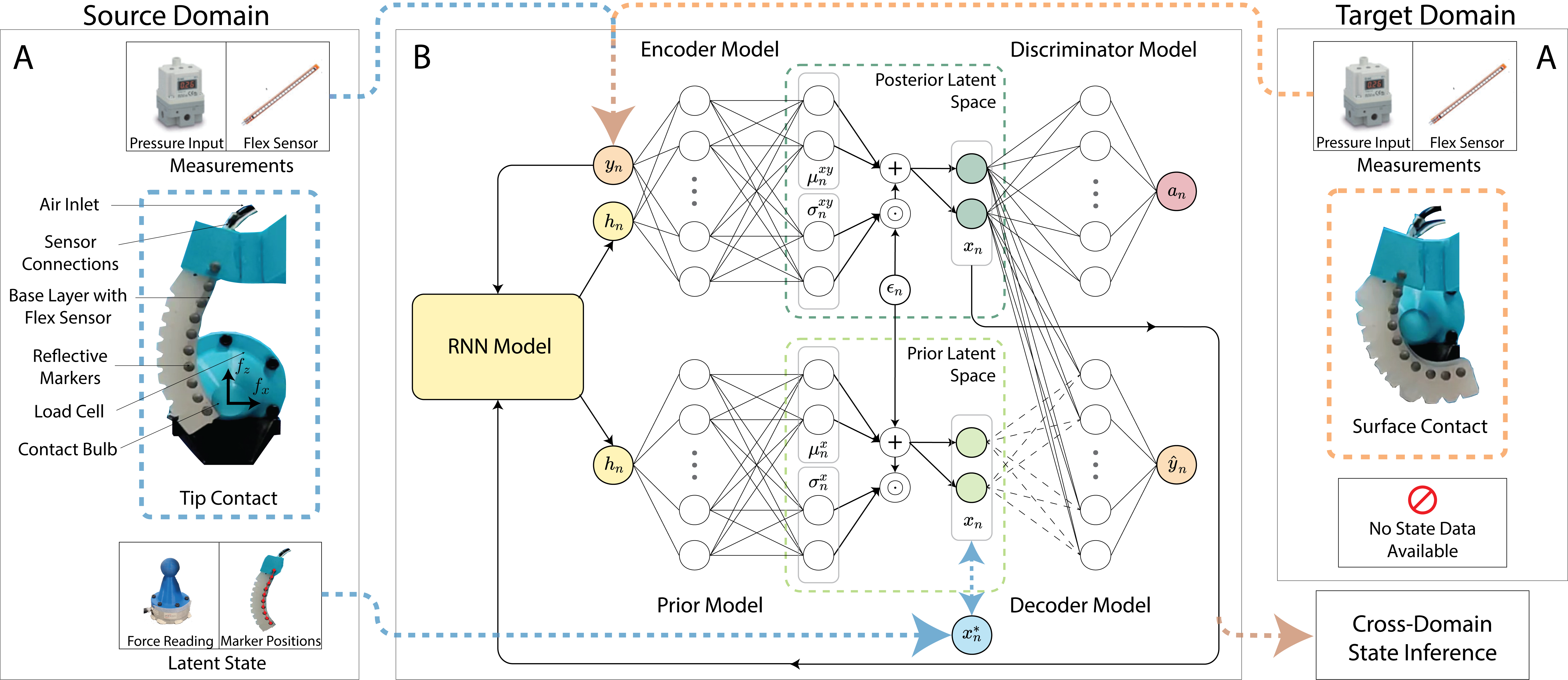}
\vspace*{-5mm}
\caption{\textbf{A cross-domain transfer learning scenario in soft robotics via the DSVB framework.} (A, Left) Source domain with contact force applied at the PSF tip (i.e., Tip Contact). (A, Right) Target domain with contact forces applied along the PSF surface (i.e., Surface Contact). $f = [f_X,f_Y]^T$ denotes the two-axis contact force.
(B) Schematic diagram of the proposed DSVB.
The measurements, $y_n$ and posterior latent state, $x_n$ are looped back to the RNN model.
The double arrow indicates semi-supervision of the partial state label $x^*_n$. Coloured dashed lines represent data flow from the source (blue) and target (orange) domains. Dashed (line) flow of the decoder model is used only for online testing. $\hat{y}_n$ denotes the reconstructed measurement.} 
\label{fig:transfer}
\vspace*{-4mm}
\end{figure*}

Recently, the deep learning paradigm of transfer learning has shown immense potential in alleviating the problem of limited labeled data. Transfer learning approaches aim to adapt the knowledge from some source domain to build data-driven prediction models of good generalization performance on a related target domain with little or no data supervision \cite{Zhuang1}. In particular, deep learning architectures based on feature encoding, such as autoencoder (AE), have been successfully applied to perform homogeneous transfer learning \cite{Jiang,Yang1,Yang2,Zhuang2,Wen}. Subsequently, Zheng et al. \cite{Chai} proposed a transfer learning framework that constructs a universal feature representation via the probabilistic counterpart of AE, variational AE (VAE), which provides a stochastic modelling solution that encapsulates predictive uncertainty. Nonetheless, \cite{Chai} only considered missing measurement inputs and assumed that the state labels are available. More importantly, these existing AE-based transfer learning approaches rely on the independent and identically distributed (i.i.d.) assumption, which falls short of capturing the rich causal structure and temporal coherence \cite{Scholkopf} of dynamic soft robot data.
To address the difficulty of acquiring representative soft robot data, in this paper, we propose a domain-adaptable sequential variational Bayes (DSVB) framework for transfer learning and state inference in soft robot systems with missing state labels on certain robot configurations (target domain). Unlike existing AE-based transfer learning approaches that construct internal feature representations without concrete semantics or tangible interpretations, our proposed framework incorporates a semi-supervised disentangle representation learning scheme to endow its latent feature space with the high-level system state representation via partial state labels on the source configurations (source domain). 
To circumvent the i.i.d. assumption, our proposed DSVB also adopts a recurrent neural network (RNN) to accurately model the nonlinear dynamics of soft robots by constructing a rich representation of the low-level temporal dependencies across the available data.
Considering the distinct dynamics and data distributions under different soft robot configurations, a feature space transfer strategy is integrated to promote domain-adaptable latent features that generalize well to both the supervised source and unsupervised target robot domains.

To the best of our knowledge, this is the first work that investigates the feasibility of a probabilistic cross-domain transfer learning framework as an alternative to the costly and laborious soft robot data collection process, alleviating the immense efforts of acquiring labelled data on many robot configurations. Our contributions are highlighted as follows:
\begin{enumerate}
  \item A DSVB framework is developed to model the complex dynamics of soft robot systems across multiple robot configurations. To facilitate representative state inference, a disentangled representation learning scheme is incorporated to semi-supervise the latent prior distribution on partial state labels from only the source domain, endowing the latent feature space with system state representation of the actual soft robot.
  \item To achieve cross-domain transfer learning amidst missing state labels on the target domain, a probabilistic feature space transfer strategy is incorporated so that the model knowledge and disentangled representation acquired from the supervised source domain can be adapted to the target domain and guide the unsupervised variational state inference.
  \item The efficacy of the proposed DSVB is validated on a pneumatic-based soft robot finger (PSF) with multiple configurations. Comparison of results with the state-of-the-art RNNs on four transfer scenarios (e.g., the one shown in Fig. \ref{fig:transfer}) demonstrates that, the DSVB achieves superior state inference performance and effective transfer learning across the supervised source and unsupervised target robot domains.
\end{enumerate}

The remainder of this paper is organized as follows:
Section II details the formulation of the proposed DSVB framework.
Section III presents a comprehensive case study on the PSF platform.
Section IV concludes the paper.

\section{Methods}
In this section, a probabilistic DSVB framework is developed to achieve a representative inference model for multi-configuration soft robots in spite of 
missing state labels. A schematic diagram of the DSVB is provided in Fig. \ref{fig:transfer}B.

\subsection{Sequential Variational Bayes Objective}

In this subsection, we introduce a general sequential Variational Bayes (VB) objective that accounts for the temporal coherence of empirical soft robot data. Variational Bayes (VB) \cite{Kingma_1} is a class of probabilistic deep learning methods that aims to maximize a lower bound of the joint measurement log-likelihood $\log p_{\theta}(Y)$. By introducing a latent state variable $X$, the variational evidence lower bound (ELBO) $\log p_{\theta}(Y) \leq \mathcal{L}^\mathrm{ELBO}(\theta,\vartheta)$ can be obtained through importance decomposition, as follows:
\begin{align} \label{eq:ELBO_general}
\begin{split}
\mathcal{L}^\mathrm{ELBO}(\theta,\vartheta) 
&= \mathbb{E}_{q_{\vartheta}(X|Y)}\left[\log \frac{p_{\theta}(Y|X) \, p_{\theta}(X)}{q_{\vartheta}(X|Y)}\right]
\end{split}
\end{align}

The latent approximate posterior distribution $q_{\vartheta}(X|Y)$, the latent prior distribution $p_{\theta}(X)$, and the measurement likelihood $p_{\theta}(Y|X)$ of the ELBO (\ref{eq:ELBO_general}) are parameterized, respectively, by the neural network (NN) models: encoder, prior, decoder, known collectively as VAE.
The parameters $\theta$ and $\vartheta$ represent the NN parameters that govern these variational models.

To capture the temporal characteristics in a given sequence $y_{\leq L}$ of measurement $y \in \mathbb{R}^{n_y}$ with sequence length $L$, 
we similarly introduce a sequence $x_{\leq L}$ of latent states $x \in \mathbb{R}^{n_x}$, and then apply ancestral factorization to the generative and (approximate) posterior distributions, respectively, to obtain
\begin{align}
\begin{split}
\label{eq:sELBO_factorization}
p_{\theta}(y_{\leq L},x_{\leq L}) &= \prod_{n=0}^{L} p_{\theta}(y_{n}|x_{\leq n},y_{< n}) \, p_{\theta}(x_{n}|x_{< n},y_{< n}) \\
q_{\vartheta}(x_{\leq L}|y_{\leq L}) &= \prod_{n=0}^{L} q_{\vartheta}(x_{n}|y_{\leq n},x_{< n})
\end{split}
\end{align}
where $n$ denotes the sample index, and $y_{\leq n}$, $x_{\leq n}$ denote the partial state and measurement sequences up to $n^{th}$ samples.

Let $Y = y_{\leq L}, X = x_{\leq L}$ and substitute (\ref{eq:sELBO_factorization}) into ELBO (\ref{eq:ELBO_general}), we obtain the following sequential ELBO (sELBO):
\begin{align} \label{eq:sELBO_general}
\begin{split}
&\; \mathcal{L}^\mathrm{sELBO}(\theta,\vartheta) \\
=&\; \mathbb{E}_{q_{\vartheta}(x_{\leq n}|y_{\leq n})}\left[\sum_{n=0}^{L} \log \frac{p_{\theta}(y_{n}|x_{\leq n},y_{< n}) \, p_{\theta}(x_{n}|x_{< n},y_{< n})}{q_{\vartheta}(x_{n}|y_{\leq n},x_{< n})}\right] \\
=&\; \sum_{n=0}^{L} \, \mathbb{E}_{q_{\vartheta}(x_{\leq n}|y_{\leq n})}\left[ \log p_{\theta}(y_{n}|x_{\leq n},y_{< n})\right] \\
&\qquad\,-\mathcal{D}^\mathrm{KL}\left[q_{\vartheta}(x_{n}|y_{\leq n},x_{< n})\|p_{\theta}(x_{n}|x_{< n},y_{< n})\right]
\end{split}
\end{align}
where $\mathcal{D}^\mathrm{KL}$ denotes the (positive-valued) Kullback–Leibler divergence (KLD). 
The conditional probabilities here encapsulate the underlying causal structure and temporal coherence of sequential data. 
This sELBO forms the basis of our proposed DSVB framework.

\subsection{Variational Recurrent Neural Network}

\begin{table*}[tb]
\small
\begin{align} \label{eq:DSVB_expand}
\begin{split}
&\mathcal{L}^\mathrm{DSVB}(\theta,\vartheta,\phi) \;=\; \mathcal{L}^\mathrm{sELBO}(\theta,\vartheta) \;+\; \mathcal{L}^\mathrm{SS}(\theta,\vartheta) \;+\; \mathcal{L}^\mathrm{BCE}(\phi) \\
&\;=\; \frac{1}{2N} \sum_{n=0}^{L} \sum_{j=1}^{n_x} \sum_{i=1}^{N} \;
\Bigg[ \, \overbrace{\log \, (2 \pi {\sigma^{y^i}_{n,j}}^{2}) + \frac{(y_n - \mu^{y^{i}}_{n,j})^{2}}{{\sigma^{y^{i}}_{n,j}}^{2}}
\;+\; \log \frac{{\sigma^{xy}_{n,j}}^{2}}{{\sigma^{x}_{n,j}}^{2}} - \frac{{\sigma^{xy}_{n,j}}^{2}}{{\sigma^{x}_{n,j}}^{2}} - \frac{(\mu^{xy}_{n,j} - \mu^{x}_{n,j})^{2}}{{\sigma^{x}_{n,j}}^{2}} + 1}^{\mathcal{L}^\mathrm{sELBO}(\theta,\vartheta)} \\
&\qquad\;+\; \underbrace{\mathbb{I}\,(x^*_n \; \text{available}) \bigg( \log \, (2 \pi {\sigma^{x^i}_{n,j}}^{2}) + \frac{(x^*_n - \mu^{x^{i}}_{n,j})^{2}}{{\sigma^{x^{i}}_{n,j}}^{2}} \, \bigg)}_{\mathcal{L}^\mathrm{SS}(\theta,\vartheta)} 
\;\underbrace{-\, \Big( \, a^i \log \varphi_{\phi}^{disc}(x^i_{n}) + (1 - a^i) \log \big(1 - \varphi_{\phi}^{disc}(x^i_{n})\big) \, \Big) \vphantom{\frac{(x^*_n - \mu^{x^{i}}_{n,j})^{2}}{{\sigma^{x^{i}}_{n,j}}^{2}}} }_{\mathcal{L}^\mathrm{BCE}(\phi)} \, \Bigg]
\end{split}
\end{align}
\vspace{-0.8cm}
\end{table*}

In this subsection, we introduce the variational RNN (VRNN) parameterization which allows stochastic gradient descent optimization of the VB objective.
First, we let the conditional latent prior distribution and measurement likelihood in (\ref{eq:sELBO_factorization}) be the following Gaussian distributions:
\begin{subequations} \label{eq:Gaussian_generative}
\begin{align}
\label{eq:Gaussian_data_likelihood}
&p_{\theta}(y_n|x_{\leq n},y_{< n}) = \mathcal{N} \Big( \mu^{y}_{n} \;,\; \Sigma^{y}_{n} \Big) \\
\label{eq:Gaussian_latent_prior}
&p_{\theta}(x_n|x_{< n},y_{< n}) = \mathcal{N} \big( \mu^{x}_{n} \;,\; \Sigma^{x}_{n} \big)
\end{align}
\end{subequations}
and let the conditional (approximate) posterior distribution in (\ref{eq:sELBO_factorization}) be the following Gaussian distribution:
\begin{align}
\label{eq:Gaussian_latent_posterior}
&q_{\vartheta}(x_n|x_{\leq n},y_{< n}) = \mathcal{N} \big( \mu^{xy}_{n} \;,\; \Sigma^{xy}_{n} \big)
\end{align}
with isotropic covariances $\Sigma^{y}_{n} = \mathrm{Diag}\big(\sigma_{n}^{y^2}\big)$, $\Sigma^{x}_{n} = \mathrm{Diag}\big(\sigma_{n}^{x^2}\big)$, $\Sigma^{xy}_{n} = \mathrm{Diag}\big(\sigma_{n}^{{xy}^2}\big)$, where $\text{Diag}(\cdot)$ denotes the diagonal function.

To allow a stochastic gradient decent optimization of the sELBO (\ref{eq:sELBO_general}), we then apply the VRNN \cite{Chung} which parameterizes the mean, standard deviation pairs: $(\mu_{n}^{y}, \sigma_{n}^{y})$, $(\mu_{n}^{x}, \sigma_{n}^{x})$ of (\ref{eq:Gaussian_generative}) and $(\mu_{n}^{xy}, \sigma_{n}^{xy})$ of (\ref{eq:Gaussian_latent_posterior}) as
\begin{subequations} \label{eq:VRNN_generative}
\begin{align}
\label{eq:VRNN_decoder}
\big[ &\mu^{y}_{n} \;,\; \sigma^{y}_{n} \big] = \varphi_{\theta}^{dec} \big(\varphi_{\theta}^{x}(x_n),h_{n}\big) \\
\label{eq:VRNN_prior}
\big[ &\mu^{x}_{n} \;,\; \sigma^{x}_{n} \big] = \varphi_{\theta}^{prior} (h_{n}) \\
\label{eq:VRNN_posterior}
\big[ &\mu^{xy}_{n} \;,\; \sigma^{xy}_{n} \big] = \varphi_{\vartheta}^{enc} \big(\varphi_{\theta}^{y}(y_n),h_{n}\big)
\end{align}
\end{subequations}
with the memory encoding hidden state
\begin{align} \label{eq:VRNN_rnn}
&h_{n} = \varphi_{\theta}^{rnn} \big(\varphi_{\theta}^{y}(y_{n-1}), \varphi_{\theta}^{x}(x_{n-1}), h_{n-1}\big).
\end{align}
Here, the decoder model $\varphi_{\theta}^{dec}$, the prior model $\varphi_{\theta}^{prior}$, the encoder model $\varphi_{\vartheta}^{enc}$, the measurement encoding model $\varphi_{\theta}^{y}$, and the state encoding model $\varphi_{\theta}^{x}$ are modelled by deep NNs (DNNs); the RNN model $\varphi_{\theta}^{rnn}$ is modelled by either the Long Short-Term Memory (LSTM) \cite{lstm} or the Gated Recurrent Units (GRU) \cite{gru} cell.

To prevent stochastic gradient with high variance \cite{Kingma_1}, the KLD $\mathcal{D}^\mathrm{KL}$ of sELBO (\ref{eq:sELBO_general}) can be analytically solved to obtain
\begin{align} \label{eq:sELBO_KLD}
\begin{split}
& - \mathcal{D}^\mathrm{KL}\left[q_{\vartheta}(x_{n}|y_{\leq n},x_{< n})\|p_{\theta}(x_{n}|x_{< n},y_{< n})\right] \\
&= \int \mathcal{N} \big( \mu^{xy}_{n} , \Sigma^{xy}_{n} \big) \log \frac{\mathcal{N} \big( \mu^{x}_{n} , \Sigma^{x}_{n} \big)}{\mathcal{N} \big( \mu^{xy}_{n} , \Sigma^{xy}_{n} \big)} \, dx \\
&= \frac{1}{2} \sum_{j=1}^{n_x} \left[ \log \frac{{\sigma^{xy}_{n,j}}^{2}}{{\sigma^{x}_{n,j}}^{2}} - \frac{{\sigma^{xy}_{n,j}}^{2}}{{\sigma^{x}_{n,j}}^{2}} - \frac{(\mu^{xy}_{n,j} - \mu^{x}_{n,j})^{2}}{{\sigma^{x}_{n,j}}^{2}} + 1 \right]
\end{split}
\end{align}
where $x_{n,j}$ denotes the $j^{th}$ element of $x_n$. Notice that the analytical solution of the KLD here is deterministic since it does not rely on Monte Carlo (MC) approximation.

Substituting (\ref{eq:sELBO_KLD}) into the sELBO (\ref{eq:sELBO_general}) and then expanding the measurement log-likelihood (\ref{eq:Gaussian_data_likelihood}), we have
\begin{align} \label{eq:sELBO_expand}
\begin{split}
& \mathcal{L}^\mathrm{sELBO}(\theta,\vartheta) \\
&= \frac{1}{2N} \sum_{n=0}^{L} \sum_{j=1}^{n_x} \sum_{i=1}^{N} \Bigg[ \, \log \, (2 \pi {\sigma^{y^i}_{n,j}}^{2}) + \frac{(y_n - \mu^{y^{i}}_{n,j})^{2}}{{\sigma^{y^{i}}_{n,j}}^{2}} \\ 
&\qquad\qquad\qquad+ \log \frac{{\sigma^{xy}_{n,j}}^{2}}{{\sigma^{x}_{n,j}}^{2}} - \frac{{\sigma^{xy}_{n,j}}^{2}}{{\sigma^{x}_{n,j}}^{2}} - \frac{(\mu^{xy}_{n,j} - \mu^{x}_{n,j})^{2}}{{\sigma^{x}_{n,j}}^{2}} + 1 \Bigg]
\end{split}
\end{align}
where $i$ denotes the particle index, $N$ denotes the number of particles. Here, we have used the MC approximation $\mathbb{E}_{q(x)} [f(x)] = \frac{1}{N} \sum_{i=1}^{N} f(x^i)$ with sampled particles $x^i \sim q(x)$ for the measurement log-likelihood expectation. The $(\mu^{y^i}_{n} , \sigma^{y^i}_{n})$ is computed via the decoder model (\ref{eq:VRNN_decoder}) using latent state particles $x_n^i \sim q_{\vartheta}(x_n|x_{\leq n},y_{< n})$, obtained using the reparameterization trick $x_n^i = \mu^{xy^i}_{n,j} + \epsilon_{n,j}^i \odot \sigma^{xy^i}_{n,j}$ where $\epsilon_{n,j}^i \sim \mathcal{N}(0, 1)$ and $(\mu^{xy^i}_{n} , \sigma^{xy^i}_{n})$ is computed via the encoder model (\ref{eq:VRNN_posterior}), and $\odot$ denotes the Hadamard 
product. 

\subsection{Disentangled Representation Learning}

In general, VRNNs rely on internal latent feature representations that do not have concrete semantics nor tangible interpretations \cite{Siddharth,Kingma_2}. One way to impose an interpretable and disentangled latent state representation is via the supervision of partial state labels.
In this section, we incorporate a semi-supervised disentangled representation learning scheme to endow the latent space of our proposed DSVB with actual system state representation.

Considering the difficulty in acquiring state labels for soft robots, this proposed scheme supervises the conditional latent prior (\ref{eq:Gaussian_latent_prior}) using partial state label $x^*$, collected only on certain robot configurations (source domain).
This state supervision on the source domain precisely specifies the axes of variation and promotes a disentangled latent representation that concurs with the actual system state representation.

In particular, we incorporate an additional log-likelihood $\mathcal{L}^\mathrm{SS}(\theta,\vartheta)$ to sELBO (\ref{eq:sELBO_expand}) which gives the following semi-supervised sELBO:
\begin{align} \label{eq:sELBO_ss_expand}
\begin{split}
& \mathcal{L}^\mathrm{sELBO}_\mathrm{SS}(\theta,\vartheta) = \mathcal{L}^\mathrm{sELBO}(\theta,\vartheta) 
+ \mathbb{I}\,(x^*_n \; \text{available}) \, \mathcal{L}^\mathrm{SS}(\theta,\vartheta) \\
&\mathcal{L}^\mathrm{SS}(\theta,\vartheta) = \sum_{n=0}^{L} \, \mathbb{E}_{q_{\vartheta}(x_{\leq n}|y_{\leq n})} \left[ \log p_{\theta}(x^*_n|x_{< n},y_{< n}) \right] \\
&= \frac{1}{2N} \sum_{n=0}^{L} \sum_{j=1}^{n_x} \sum_{i=1}^{N} \, \bigg[ \log \, (2 \pi {\sigma^{x^i}_{n,j}}^{2}) + \frac{(x^*_n - \mu^{x^{i}}_{n,j})^{2}}{{\sigma^{x^{i}}_{n,j}}^{2}} \, \bigg]
\end{split}
\end{align}
where $\mathbb{I}\,(\cdot)$ denotes the indicator function.
Similarly, here we have expanded the latent prior (\ref{eq:Gaussian_latent_prior}) in $\mathcal{L}^\mathrm{SS}(\theta,\vartheta)$ and used the MC approximation for its expectation. The $(\mu^{x^i}_{n} , \sigma^{x^i}_{n})$ is computed via the prior model (\ref{eq:VRNN_prior}) using the hidden state particles $h_n^i$, output by the RNN model (\ref{eq:VRNN_rnn}).

Based on this semi-supervised sELBO (\ref{eq:sELBO_ss_expand}), the latent prior (\ref{eq:Gaussian_latent_prior}) is partially conditioned on state labels $x_n^*$, thus endowing it with the disentangled state representation of the actual system. The latent (approximate) posterior (\ref{eq:Gaussian_latent_posterior}) then inherits this state representation via the KLD (\ref{eq:sELBO_KLD}), which minimizes the distance between the latent prior and (approximate) posterior.

\subsection{Probabilistic Domain Adversarial Training}

In practice, it is common for complex soft robot systems under different robot configurations to exhibit distinct dynamics and data distributions. Additionally, the disentangled representation learning described in the previous section can only be applied to the source domain, where state labels are available. However, the lack of state labels in the target domain can lead to an unconditioned latent space, potentially resulting in entangled representations that do not accurately reflect the actual system state representation. This limitation can negatively impact state inference performance.

Taking these into account, in this subsection, we incorporate a probabilistic latent space transfer strategy to encourage domain-adaptable latent features. This transfer strategy facilitate the adaptation of the disentangled representation constructed for the supervised source domain (using partial state labels), to the latent space on the unsupervised target domain. Thus, our objective is to minimize the discrepancy between the latent distributions of the source and target domains. To achieve this, we introduce a probabilistic binary cross entropy (BCE) loss, given by:
\begin{align} \label{eq:DAT}
\begin{split}
\mathcal{L}^\mathrm{BCE}(\phi) = \frac{1}{N} \sum_{i=1}^{N} & - \Big( \, a^i \log \varphi_{\phi}^{disc}(x^i_{n}) \\
&+ (1 - a^i) \log \, \big(1 - \varphi_{\phi}^{disc}(x^i_{n})\big) \, \Big)
\end{split}
\end{align}
where $a^i \in \{0, 1\}$ is the domain label which indicates if the latent sample $x^i_{n}$ belongs to source or target domain, and $\varphi_{\phi}^{disc}$ is a
NN-parameterized discriminator model that predicts the domain label given the latent sample.
Incorporating (\ref{eq:DAT}) to the semi-supervised ELBO (\ref{eq:sELBO_ss_expand}), we obtain the final DSVB loss $\mathcal{L}^\mathrm{DSVB}$ as 
equation (\ref{eq:DSVB_expand}). 

Motivated by the deterministic domain adversarial training (DAT) strategy \cite{Ganin}, we optimize the parameters $(\theta,\vartheta,\phi)$ in an adversarial fashion as follows:
\begin{align} \label{eq:DSVB_objective}
\begin{split}
(\theta,\vartheta) &= \argmax_{\theta,\vartheta} \, \mathcal{L}^\mathrm{DSVB}(\theta,\vartheta,\phi) \\
\phi &= \argmin_{\phi} \, \mathcal{L}^\mathrm{DSVB}(\theta,\vartheta,\phi)
\end{split}
\end{align}
Under this training strategy, 
the VRNN parameters $(\theta,\vartheta)$ are optimized to generate latent samples $z^i$ that fool the classifier $\varphi_{\phi}^{disc}$, making the corresponding domain label $a^i$ unrecognizable.
At the same time,
the discriminator model parameter $\phi$ is optimized to accurately distinguish the domain label of the generated latent sample. Such an adversarial competition is expected to achieve the Nash equilibrium that generalizes the latent space to both the source and target domains \cite{Chai}.

\section{Case Study}

\subsection{Soft Robot System Description}
In this work, we adopt a pneumatic-based soft finger (PSF) as shown in Fig. \ref{fig:transfer}A to be used in our case study. This PSF comprises a popular class of soft actuators known as PneuNet \cite{Soft_robots_chemists}, commonly used in a soft gripper to pick up odd-shaped and fragile objects which is challenging for conventional rigid grippers \cite{surya1}. The PSF consists of two sections: the main body with a series of pleated pneumatic chambers and an in-extensible base layer embedded with a resistive flex sensor (Spectra Symbol).

During the experiment, the PSF is actuated using a series of pseudo-random pressure levels within a predetermined pressure range. Two different actuation patterns, namely, \textit{Oscillatory Actuation} and \textit{Random Actuation} are used to verify the generality of our approach to different input signals. In the former, the PSF is actuated using a gradual oscillatory input pressure, while the latter uses a faster random pressure actuation pattern.
For each actuation pattern, two experimental configurations were captured. In the first configuration, \textit{Tip Contact}, the PSF is actuated and exposed to a contact bulb placed close to its tip, as shown in Fig. \ref{fig:transfer}A (Left). In the second configuration, \textit{Surface Contact}, the PSF is configured to randomly move along the $Z$-axis with its entire front surface exposed to the contact bulb, as shown in Fig. \ref{fig:transfer}A (Right). These configurations simulate object grasping at arbitrary locations along the PSF.

\begin{figure}[tb]
\centering
\includegraphics[width=0.8\columnwidth]{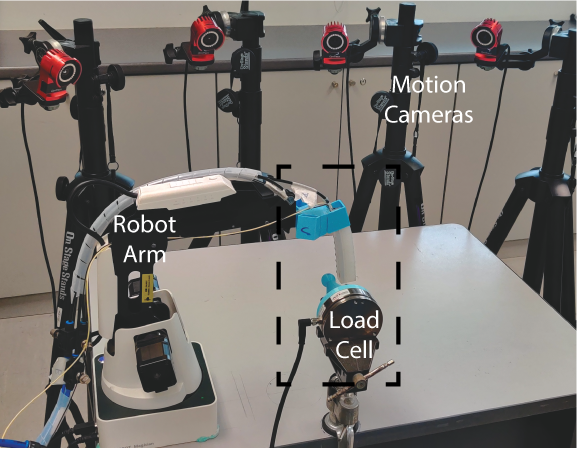}
\vspace*{-2mm}
\caption{\textbf{The setup of pneumatic-based soft finger (PSF)}. 
Four motion cameras are used to track the markers attached on the PSF. A rigid robot arm is used to maneuver the position of PSF.
}
\label{fig:scenarios}
\vspace{-2mm}
\end{figure}

During data collection, 10 reflective camera markers are placed evenly along the inextensible base layer to capture the PSF motion. The marker positions are recorded by three motion cameras (OptiTrack Flex13, NaturalPoint Inc.) as 2D Cartesian coordinates, as shown in Fig. \ref{fig:scenarios}. 
The contact forces are measured along axes $X,Z$ using a load cell (Axia80, ATI Industrial Automation Inc.) attached to the contact bulb, as depicted in Fig. \ref{fig:scenarios}.
Data is sampled at 10Hz, and each sample includes the actuation pressure, flex sensor reading, marker positions and contact forces. On each actuation pattern and experimental configuration, we sample 15000 data points for training and 3000 data points for testing. 

\subsection{Experimental Setup}

To assess the domain-adaptability of the developed DSVB framework under missing state labels, we supervise the latent state using only state labels from the source domain. The performance of the DSVB is then evaluated through state inference on data from the target domain only. In this work, we design four transfer scenarios with different source and target domains composed of permutations of the captured experimental configurations and actuation patterns. Concretely, these are \textit{Tip Contact} as the source domain with \textit{Surface Contact} as the target domain (Scenario 1) and vice versa (Scenario 2). Alternatively, we also have \textit{Oscillatory Actuation} as the source domain with \textit{Random Actuation} as the target domain (Scenario 3) and vice versa (Scenario 4). By training and testing on differing source and target domains, we demonstrate the ability of the proposed method in transfer learning the well learned representations from the supervised source domain to the unseen target domain. We benchmark the proposed DSVB against two state-of-the-art RNNs, Gated Recurrent Unit (GRU) and Long Short-Term Memory (LSTM) networks. In contrast to the proposed method, these models are deterministic and are trained using only the partial measurement data and state label pairs of the source domain.

In this case study, we validate two implementations of the DSVB, where the RNN model $\varphi_{\theta}^{rnn}$ is either parameterized by a single-layer vanilla GRU \cite{gru}, namely DSVB-GRU, or a single-layer LSTM \cite{lstm}, DSVB-LSTM. In both cases, the RNN model uses a hidden state $h_n \in \mathbb{R}^{128}$ of size 128. The prior model $\varphi_{\theta}^{prior}$, encoder model $\varphi_{\theta}^{enc}$ and the decoder model $\varphi_{\theta}^{dec}$ are feed-forward NNs (FNNs) with hidden layer architectures \{128\}, \{128, 128\} and \{128, 64\}, respectively, where each entry of the curly brackets is a hidden layer and the value denotes layer width.
The state and measurement encoding models $\varphi_{\theta}^{x}$ and $\varphi_{\theta}^{y}$, respectively, are FNNs with hidden layer architectures \{32\} and \{22\}.
The domain classifier $\varphi_{\phi}^{disc}$ is parameterized by a single GRU layer with 128 hidden states and a succeeding FNN with hidden layer architecture \{128, 1\}. For the benchmark methods, we consider a vanilla GRU or LSTM with 128 hidden states, followed by a FNN \{128\} that maps the hidden states to the system states. The NN parameters across all these comparison methods are initiated with the same values. 

All the comparison methods (DSVB and benchmark methods) are trained on a sequence length of 100 and a batch size of 128, using the Adam optimizer with a learning rate of 0.001 for 50 epochs to allow the methods to converge. During inference time, the comparison methods output the estimated latent state labels (contact forces and marker positions) using the original measurement data (flex reading and pressure input) from the test set. The latent state labels in the test set serve as ground truth for result validation.

\begin{table*}[tb]
\centering
\caption{Comparison Between State Inference Results of the Proposed DSVB and Benchmark State-of-the-Art Recurrent Neural Networks on Source and Target Domains}
\label{tab:rmse}
\vspace*{-2mm}
\resizebox{\textwidth}{!}{%
\begin{tabular}{@{}lccccccccccc@{}}
\toprule
\multicolumn{1}{c}{\multirow{2}{*}{Method}} &
  \multicolumn{2}{c}{Scenario 1} &
  \multicolumn{1}{l}{} &
  \multicolumn{2}{c}{Scenario 2} &
  \multicolumn{1}{l}{} &
  \multicolumn{2}{c}{Scenario 3} &
  \multicolumn{1}{l}{} &
  \multicolumn{2}{c}{Scenario 4} \\ \cmidrule(lr){2-3} \cmidrule(lr){5-6} \cmidrule(lr){8-9} \cmidrule(l){11-12} 
\multicolumn{1}{c}{} &
  \begin{tabular}[c]{@{}c@{}}Source \\ (Tip Contact)\end{tabular} &
  \begin{tabular}[c]{@{}c@{}}Target\\ (Surface Contact)\end{tabular} &
  \multicolumn{1}{l}{} &
  \begin{tabular}[c]{@{}c@{}}Source\\ (Surface Contact)\end{tabular} &
  \begin{tabular}[c]{@{}c@{}}Target\\ (Tip Contact)\end{tabular} &
  \multicolumn{1}{l}{} &
  \begin{tabular}[c]{@{}c@{}}Source \\ (Osc. Actuation)\end{tabular} &
  \begin{tabular}[c]{@{}c@{}}Target\\ (Rand. Actuation)\end{tabular} &
  \multicolumn{1}{l}{} &
  \begin{tabular}[c]{@{}c@{}}Source\\ (Rand. Actuation)\end{tabular} &
  \begin{tabular}[c]{@{}c@{}}Target\\ (Osc. Actuation)\end{tabular} \\ \midrule
LSTM &
  \textbf{0.952±0.024} &
  2.678±0.074 &
   &
  \textbf{0.448±0.014} &
  1.087±0.026 &
   &
  \textbf{0.939±0.032} &
  1.564±0.028 &
   &
  \textbf{0.883±0.009} &
  1.522±0.022 \\
DSVB-LSTM &
  1.237±0.053 &
  \textbf{0.961±0.017} &
   &
  0.594±0.034 &
  \textbf{0.680±0.049} &
   &
  1.187±0.021 &
  \textbf{1.005±0.014} &
   &
  0.925±0.046 &
  \textbf{1.058±0.038} \\ \midrule
GRU &
  \textbf{0.939±0.024} &
  2.863±0.065 &
   &
  \textbf{0.442±0.013} &
  1.096±0.012 &
   &
  \textbf{0.915±0.031} &
  1.577±0.040 &
   &
  0.890±0.022 &
  1.487±0.020 \\
DSVB-GRU &
  1.179±0.009 &
  \textbf{0.940±0.025} &
   &
  0.551±0.017 &
  \textbf{0.673±0.049} &
   &
  1.157±0.069 &
  \textbf{0.986±0.017} &
   &
  \textbf{0.886±0.026} &
  \textbf{1.021±0.044} \\ \bottomrule
\end{tabular}%
}
\end{table*}

\begin{figure*}[tb]
  \centering
  \includegraphics[width=\textwidth]{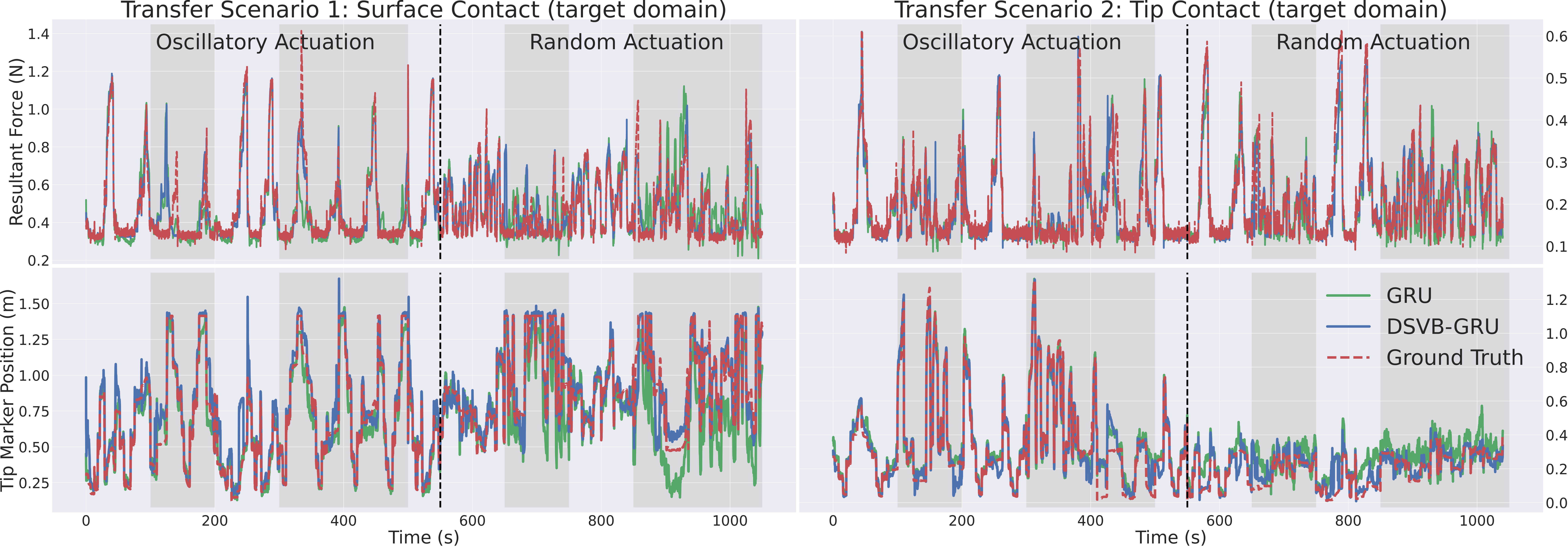}
  \vspace*{-7mm}
  \caption{Normalized time-series state estimation of the best benchmark (GRU) and proposed method (DSVB-GRU) on the target domains of transfer Scenario 1 (left) and Scenario 2 (right). Resultant force (top) is the 2-norm of contact forces $X,Z$ and tip marker position (bottom) is the 2-norm of tip marker coordinates $X,Z$. The shaded and unshaded regions indicate time windows with states from the (unsupervised) target domain and (supervised) source domain, respectively.}
  \label{fig:estimations}
  \vspace*{-4mm}
\end{figure*}

\begin{figure}[tb]
  \centering
  \includegraphics[width=\columnwidth]{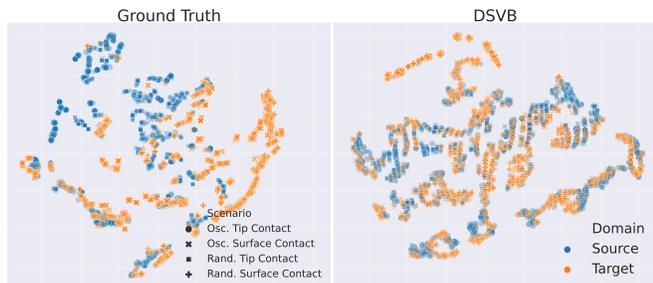}
  \vspace*{-6mm}
  \caption{Comparision of tSNE visualizations of source and target domains for the ground truth and the posterior latent states in the proposed DSVB.}
  \label{fig:tsne}
\end{figure}

\subsection{Results \& Discussions}
Table \ref{tab:rmse} presents the prediction performance (root-mean-squared errors) of the four scenarios described, where the means and standard deviations are aggregated over five runs. For an ablative study, we compare the RNN benchmark models (either GRU or LSTM) directly with the corresponding DSVB implementations.
The results show that our proposed DSVB framework achieves comparable performances to the benchmark methods when evaluated on the supervised source domain. However, the differences in state estimation error between the proposed DSVB and the benchmark methods become much more significant on the unsupervised target domain, which contains system dynamics that were not directly exposed to the models during training. Figure \ref{fig:estimations} illustrates that the GRU has difficulty producing accurate estimates on the target domains of transfer scenarios 1 and 2, presumably due to the distinct dynamics in the source and target domains. In contrast, the DSVB yields consistent results across both domains, indicating that the proposed framework achieves effective transfer learning.

Furthermore, to verify the capability of the probabilistic domain adversarial training scheme, Figure \ref{fig:tsne} shows the tSNE distribution for the source and target domains from the ground truth and the proposed DSVB method for Scenario 1. Here, we observe that the tSNE embeddings of the proposed method exhibit smaller discrepancies between the two domains compared to the ground truth. The affinity of the source and target domains highlights the capability of the DSVB framework in transferring the disentangled representations acquired from the supervised source domain to achieve domain-adaptable latent features which facilitate cross-domain transfer learning even in the absence of missing state labels on the target domain.

\section{Conclusion}
The capability of DSVB to perform 
transfer learning offers a solution for accurate soft robot modelling and inference in the absence of adequate state labels. The experimental results show %
that labelled data only need to be collected on relatively simple source robot configurations, and the knowledge learnt can be transferred by the proposed framework to multiple unlabelled target configurations. For future works, we could investigate the feasibility of DSVB on more involved non-planar soft robots \cite{nakajima,bruder,meerbeek}. We plan to also incorporate normalizing flows \cite{Rezende_2,Kingma_3} to our proposed DSVB.

                                  
\clearpage
\bibliography{references}
\bibliographystyle{IEEEtran}

\end{document}